\documentstyle[aaai,epsf]{article}

\def\jape{{\sf J\kern-.15em\raise-.3ex\hbox{\footnotesize\sf
A}\kern-.15em\raise.5ex\hbox{\footnotesize\sf
P}\kern-.15em\kern.120emE\kern.120em-\kern.120em1}}

\begin{document}
\epsfverbosetrue

%
%

\title{An implemented model of punning riddles}

\author{Kim Binsted\thanks{Thanks
are due to Canada Student Loans, the Overseas Research Students Scheme,
and the St Andrew's Society of Washington, DC, for their financial
support.} and Graeme Ritchie\\
	Department of Artificial Intelligence\\
	University of Edinburgh\\
	Edinburgh, Scotland EH1 1HN\\
	kimb@aisb.ed.ac.uk graeme@aisb.ed.ac.uk\\}
\maketitle
\begin{abstract}
\begin{quote}

In this paper, we discuss a model of simple question--answer punning,
implemented in a program, \jape, which generates riddles from
humour--independent lexical entries. The model uses two main types of
structure: {\em schemata}, which determine the relationships between key
words in a joke, and {\em templates}, which produce the surface form of
the joke. \jape\ succeeds in generating pieces of text that are
recognizably jokes, but some of them are not very good jokes. We mention
some potential improvements and extensions, including post--production
heuristics for ordering the jokes according to quality.

\end{quote}
\end{abstract}

\section{Humour and artificial intelligence}
\label{introduction}

If a suitable goal for AI research is to get a computer to do ``\ldots a
task which, if done by a human, requires intelligence to perform,''
\cite{Minsky2}, then the production of humorous texts, including jokes
and riddles, is a fit topic for AI research. As well as probing some
intriguing aspects of the notion of ``intelligence'', it has the
methodological advantage (unlike, say, computer art) of leading to more
directly falsifiable theories: the resulting humorous artefacts can be
tested on human subjects.

Although no computationally tractable model of humour as a whole has yet
been developed (see \cite{AR1} for a general theory of verbal humour,
and \cite{AttardoNew} for a comprehensive survey), we believe that by
tackling a very limited and linguistically-based set of phenomena, it is
realistic to start developing a formal symbolic account.

One very common form of humour is the question-answer joke, or riddle.
Most of these jokes (e.g. almost a third of the riddles in the
Crack-a-Joke Book \cite{CAJB}) are based on some form of pun. For
example:
\begin{verse}

What do you use to flatten a ghost? {\em A spirit level.} \cite{CAJB}

\end{verse}
This riddle is of a general sort which is of particular interest for a
number of reasons. The linguistics of riddles has been investigated
before (e.g. \cite{PG}). Also, there is a large corpus of riddles to
examine: books such as \cite{CAJB} record them by the thousand. Finally,
riddles exhibit more regular structures and mechanisms than some other
forms of humour.

We have devised a formal model of the punning mechanisms underlying some
subclasses of riddle, and have implemented a computer program which uses
these symbolic rules and structures to construct punning riddles from a
humour-independent (i.e. linguistically general) lexicon. An informal
evaluation of the performance of this program suggests that its output
is not significantly worse than that produced by human composers of such
riddles.

\section{Punning riddles}
\label{jokes-tackled}

Pepicello and Green \cite{PG} describe the various strategies
incorporated in riddles. They hold the common view that humour is
closely related to ambiguity, whether it be linguistic (such as the
phonological ambiguity in a punning riddle) or contextual (such as
riddles that manipulate social conventions to confuse the listener).
What the linguistic strategies have in common is that they ask the
``riddlee'' to accept a similarity on a phonological, morphological, or
syntactic level as a point of {\em semantic} comparison, and thus get
fooled (cf. ``iconism'' \cite{AttardoNew}). Riddles of this type
are known as {\em puns}.

We decided to select a subset of riddles which displayed regularities at
the level of semantic, or logical, structure, and whose structures could
be described in fairly conventional linguistic terms (simple lexical
relations). As a sample of existing riddles, we studied  ``The
Crack-a-Joke Book'' \cite{CAJB}, a collection of jokes chosen by British
children. These riddles are simple, and their humour generally arises
from their punning nature, rather than their subject matter. This sample
does not represent sophisticated adult humour, but it suffices for an
initial exploration.

\label{tax}

There are three main strategies used in puns to exploit phonological
ambiguity: {\em syllable substitution}, {\em word substitution}, and
{\em metathesis}. This is not to say that other strategies do not exist;
however, none were found among the large number of punning jokes
examined.

\paragraph{Syllable substitution:}
Puns using this strategy confuse a syllable (or syllables) in a word
with a similar- or identical-sounding word. For example:
\begin{quote}
What do short-sighted ghosts wear? {\em Spooktacles.} \cite{CAJB}
\end{quote}

\paragraph{Word substitution:}
Word substitution is very similar to syllable substitution. In this
strategy, an entire word is confused with another similar- or
identical-sounding word. For example:
\begin{quote}
How do you make gold soup? {\em Put fourteen carrots in it.} \cite{CAJB}
\end{quote}

\paragraph{Metathesis:}
Metathesis is quite different from syllable or word substitution. Also
known as {\em spoonerism}, it uses a reversal of sounds and words to
suggest (wrongly) a similarity in meaning between two
semantically-distinct phrases. For example:
\begin{quote}
What's the difference between a very short witch and a deer running from
hunters? {\em One's a stunted hag and the other's a hunted stag.}
\cite{CAJB}
\end{quote}

All three of the above-described types of pun are potentially tractable
for detailed formalisation and hence computer generation.  We chose to
generate only word-substitution puns, simply because lists of
phonologically identical words ({\em homonyms}) are readily available,
whereas the other two types require some kind of sub-word comparison.
In particular, the class of jokes which we chose to generate all: use
word substitution; have the substituted word in the {\em
punchline} of the joke, rather than the question; and substitute a
homonym for a word in a {\em common noun phrase} (cf. the
``spirit level'' riddle cited earlier).
These restrictions are simply to reduce the scope of the research even
further, so that the chosen subset of jokes can be covered in a
comprehensive, rigorous manner. We believe that our basic model,
with some straightforward extensions, is general enough to cover other
forms.

\section{Symbolic descriptions}

\label{theory}

\begin{figure}[htb]
\epsfxsize = 0.45\textwidth
\epsfbox{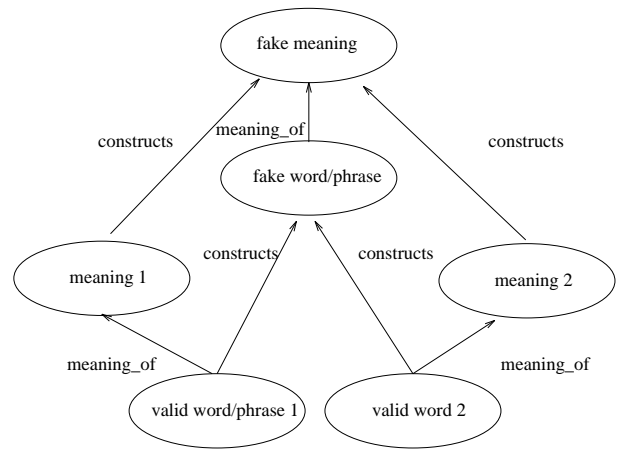}
\caption{The relationships between parts of a pun}
\label{theoryfig}
\end{figure}

Our analysis of word-substitution riddles is based (semi-formally)
on the following essential items, related as shown in
Figure~\ref{theoryfig}:

\begin{quote}
$\bullet$ a valid English word/phrase\\
$\bullet$ the meaning of the word/phrase\\
$\bullet$ a shorter word, phonologically similar to part of the
word/phrase\\
$\bullet$ the meaning of the shorter word\\
$\bullet$ a fake word/phrase, made by substituting the shorter
word into the word/phrase\\
$\bullet$ the meaning of the fake word/phrase, made by combining the
meanings of the original word/phrase and the shorter word.
\end{quote}

At this point, it is important to distinguish between the mechanism for
building the {\em meaning} of the fake word/phrase, and the mechanism
that uses that meaning to build a question with the word/phrase as an
answer. Consider the joke:
\begin{quote}
What do you give an elephant that's exhausted? {\em Trunkquillizers.}
\cite{CAJB}
\end{quote}

In this joke, the word ``trunk'', which is phonologically similar to the
syllable ``tranq'', is substituted into the valid English word
``tranquillizer''. The resulting fake word ``trunkquillizer'' is given a
meaning, referred to in the question part of the riddle, which is some
combination of the meanings of ``trunk'' and ``tranquillizer'' (in this
case, a tranquillizer for elephants). The following questions use the
same meaning for `trunkquillizer', but refer to that meaning in
different ways:
\begin{quote}
$\bullet$ What do you use to sedate an elephant?\\
$\bullet$ What do you call elephant sedatives?\\
$\bullet$ What kind of medicine do you give to a stressed-out
elephant?
\end{quote}
On the other hand, {\em these} questions are all put together in the
same way, but from different constructed meanings:
\begin{quote}
$\bullet$ What do you use to sedate an elephant?\\
$\bullet$ What do you use to sedate a piece of luggage?\\
$\bullet$ What do you use to medicate a nose?
\end{quote}

We have adopted the term {\em schema} for the symbolic description of
the underlying configuration of meanings and words,  and {\em template}
for the textual patterns used to construct a question-answer pair.

\subsection{Lexicon}

Our minimal assumptions about the structure of the lexicon are as
follows. There is a (finite) set of {\em lexemes}. A lexeme is an
abstract entity, roughly corresponding to a meaning of a word or phrase.
Each lexeme has exactly one entry in the lexicon, so if a word has two
meanings, it will have two corresponding lexemes. Each lexeme may have
some {\em properties} which are true of it (e.g. being a noun), and
there are a number of possible {\em relations} which may hold between
lexemes (e.g. synonym,  homonym, subclass). Each lexeme is also
associated with a {\em near-surface form} which indicates (roughly) the
written form of the word or phrase.

\subsection{Schemata}

A {\em schema} stipulates a set of relationships which must hold between
the lexemes used to build a joke. More specifically, a schema determines
how real words/phrases are glued together to make a fake word/phrase,
and which parts of the lexical entries for real words/phrases are used
to construct the meaning of the fake word/phrase.

\begin{figure}[htb]
\epsfxsize = 0.45\textwidth
\epsfbox{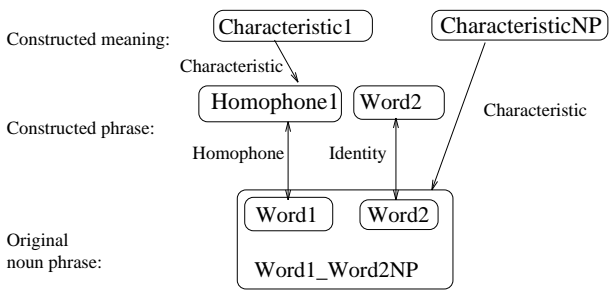}
\caption{The {\em lotus} schema}
\label{lotusfig}
\end{figure}

There are many different possible schemata (with obscure symbolic
labels which the reader can ignore). For
example, the schema in Figure~\ref{lotusfig} constructs a fake phrase by
substituting a homonym for the first word in a real phrase, then builds
its meaning from the meaning of the homonym and the real phrase.

The schema shown in Figure~\ref{lotusfig} is {\em uninstantiated};
that is, the actual lexemes to use have not yet been specified.
Moreover, some of the relationships are still quite general --- the
{\em characteristic} link merely indicates that {\em some} lexical
relationship must be present, and the {\em homonym} link allows
either a homophone or the same word with an alternative
meaning. Instantiating a schema means inserting
lexemes in the schema, and specifying the exact relationships between
those lexemes (i.e. making exact the {\em characteristic} links).  For
example, in the lexicon, the lexeme {\bf spring\_cabbage} might
participate in relations as follows:

\small
\begin{quote}
\begin{verbatim}
class(spring_cabbage, vegetable)
location(spring_cabbage, garden)
action(spring_cabbage, grows)
adjective(spring_cabbage, green)
....
\end{verbatim}
\end{quote}
\normalsize

If {\bf spring\_cabbage} were to be included in a schema, at one end of
a {\em characteristic} link, the other end of the link could be
associated with any one, or any combination of, these values (vegetable,
garden, etc), depending on the exact label (class, location, etc.)
chosen for the characteristic link.

\begin{figure}[htb]
\epsfxsize = 0.45\textwidth
\epsfbox{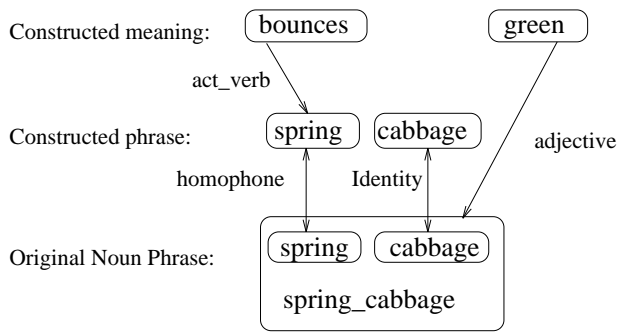}
\caption{A completely instantiated {\em lotus} schema}
\label{filled}
\end{figure}

The completely instantiated {\em lotus} schema in Figure~\ref{filled}
could (with an appropriate template --- see below) be
used to construct the joke:

\begin{quote}
What's green and bounces? {\em A spring cabbage.} \cite{CAJB}
\end{quote}

\subsection{Templates}
\label{tem}

A template is used to produce the surface form of a joke from the
lexemes and relationships specified in an instantiated schema.
Templates are not inherently humour-related. Given a (real or
nonsense) noun phrase, and a meaning for that noun phrase (genuine or
constructed), a template builds a suitable question-answer pair.
Because of the need to provide a suitable amount of information in
the riddle question, every schema has to be associated with a set of
appropriate templates. Notice that the precise choice of relations for
the
under-specified ``characteristic'' links will also affect the
appropriateness of a template.  (Conversely, one could say that the
choice of template influences the choice of lexical relation for the
characteristic link, and this is in fact how we have implemented it.)
Abstractly, a template is a mechanism which maps a
set of lexemes (from the instantiated schema) to the surface form
of a joke.

\section{The JAPE-1 computer program}
\label{program}
\subsection{Introduction}

We have implemented the model described earlier in a computer program
called \jape, which produces the chosen subtype of jokes --- riddles
that use homonym substitution and have a noun phrase punchline. Such
riddles are representative of punning riddles in general, and include
approximately one quarter of the punning riddles in \cite{CAJB}.

\jape\ is significantly different from other attempts to computationally
generate humour in various ways: its lexicon is humour-independent
(i.e. the structures that generate the riddles are distinct from the
semantic and syntactic data they manipulate), and  it generates riddles
that are similar on a strategic and structural level, rather than in
surface form.

\label{top}

\jape's main mechanism attempts to construct a punning riddle based on a
common noun phrase. It has several distinct knowledge bases with which
to accomplish this task: the lexicon (including the homonym base), a set
of
schemata, a set of templates, and a post-production checker.

\subsection{Lexicon}
\label{lexicon}

The lexicon contains humour--independent semantic and syntactic
information about the words and noun phrases entered in it, in the
form of ``slots'' which can contain other lexemes or may contain
other symbols. A typical entry might be:
{\small \begin{verbatim}
lexeme = jumper_1            countable = yes
category = noun              class = clothing
written_form = ``jumper''    specifying_adj = warm
vowel_start = no             synonym =  sweater
\end{verbatim}}

Although the lexicon stores syntactic information, the amount of syntax
used by the rest of the program is minimal. Because the templates are
based on certain fixed forms, the only necessary syntactic information
has
to do with the syntactic category, verb person, and determiner
agreement. Also, the lexicon need only contain entries for nouns, verbs,
adjectives, and common noun phrases --- other types of word
(conjunctions, determiners, etc) are built into the templates.
Moreover, because the model implemented in \jape\ is restricted to
covering riddles with noun phrase punchlines, the schemata require
{\em semantic} information only for nouns and adjectives.

\label{homonyms}
The ``homonym'' relation between lexemes was implemented as a
separate {\em homonym base} derived from a list \cite{Townsend}
of homophones in American English, shortened considerably for
our purposes. The list now contains only common, concrete nouns and
adjectives. The homonym base also includes words with two distinct
meanings (e.g. ``lemon'', the fruit, and ``lemon'', slang for a
low-quality car).

\subsection{Schemata}
\label{schemata}

\begin{figure}[htb]
\epsfxsize = 0.45\textwidth
\epsfbox{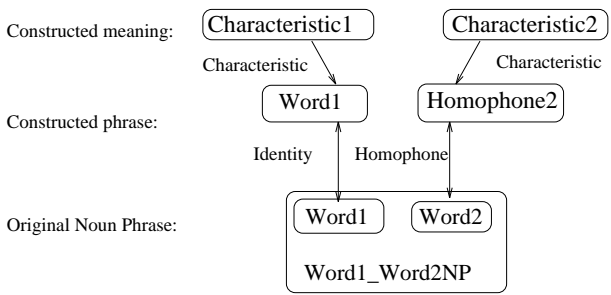}
\caption{The uninstantiated {\em jumper} schema}
\label{jumper}
\end{figure}

\begin{figure}[htb]
\epsfxsize = 0.45\textwidth
\epsfbox{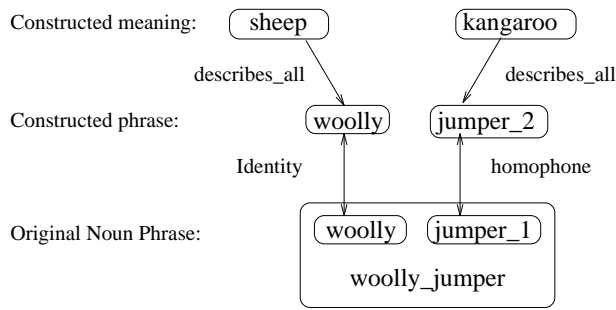}
\caption{The instantiated {\em jumper} schema, with links suitable
for the {\em syn\_syn} template. Gives the riddle: What do you get
 when you cross a sheep and a kangaroo? {\em A woolly jumper.}}
\label{jumpfill}
\end{figure}

\begin{figure}[htb]
\epsfxsize = 0.45\textwidth
\epsfbox{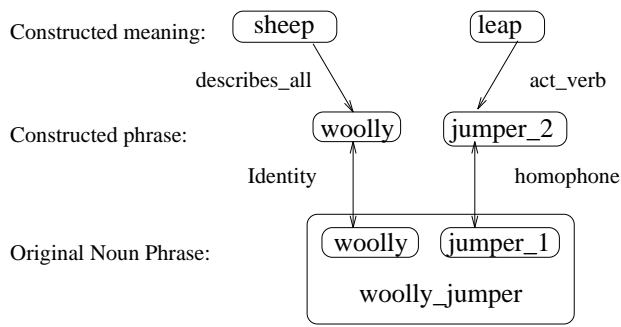}
\caption{The instantiated {\em jumper} schema, with links suitable for
the {\em syn\_verb} template. Gives the riddle:
What do you call a sheep that can leap? {\em A woolly jumper.}}
\label{jumpfill2}
\end{figure}

\jape\ has a set of six schemata, one of which is the {\em jumper}
schema, shown in Figure~\ref{jumper}. The same schema, instantiated in
two different ways, is shown in Figure~\ref{jumpfill} and
Figure~\ref{jumpfill2}.

\subsection{Templates}
\label{templates}

Since riddles often use certain fixed forms (for example, ``What do you
get when you cross  \_\_\_ with \_\_\_ ?''), \jape's templates embody
such standard forms. A \jape\ template consists of some fragments of
canned text with ``slots'' where generated words or phrases can be
inserted, derived from the lexemes in an instantiated schema. For
example, the {\em syn\_syn} template:

\begin{quote}

What do you get when you cross {\bf [text fragment generated from the
first
characteristic lexeme(s)]} with {\bf [text fragment generated from the
second
characteristic lexeme(s)]}? {\em [the constructed noun phrase].}

\end{quote}

A template also specifies the values it requires to be used for
``characteristic'' links in the schema; the {\em describes\_all}
labels in  Figure~\ref{jumpfill} are derived from the {\em syn\_syn}
template.  When the schema has
been fully instantiated, \jape\  selects one of the associated
templates, generates text fragments from the lexemes, and slots those
fragments into the template.

Another template which can be used with the {\em jumper} schema (see
Figure~\ref{jumpfill2}) is the {\em syn\_verb} template:
\begin{quote}
What do you call {\bf [text fragment generated from the first
characteristic
lexeme(s)]} that {\bf [text fragment generated from the second
characteristic
lexeme(s)]}? {\em [the constructed noun phrase.]}
\end{quote}

%

\subsection{Post-production checking}
\label{checker}

To improve the standard of the jokes slightly,
some simple checks are made on the final form. The first is that none of
the lexemes used to build the question and punchline are accidentally
identical; the second  is that the lexemes used to build the nonsense
noun phrase and its meaning, do not build a {\em genuine} common
noun phrase.

\section{The evaluation procedure}
\label{methodology}
\label{evaluation}

An informal evaluation of \jape\ was carried out, with three stages:
{\em data acquisition}, {\em common knowledge judging} and {\em joke
judging}. During the data acquisition stage, volunteers unfamiliar with
\jape\ were asked to make lexical entries for a set of words given to
them. These definitions were then  sifted by a ``common knowledge
judge'' (simply to check for errors and excessively obscure
suggestions),
entered into \jape's lexicon, and a substantial set of jokes
were produced.  A different group of volunteers then gave verdicts, both
quantitative and qualititative, on these jokes.
The use of volunteers to write lexical entries was a way of
making the testing slightly more rigorous. We did not have access
to a suitable large lexicon, but if we had hand-crafted the
entries ourselves there would have been the risk of bias (i.e.
humour-oriented information) creeping in.

\jape\  produced a set of 188 jokes in near-surface form, which were
distributed in batches to 14 judges, who gave the jokes scores on a
scale from 0 (``Not a joke. Doesn't make any sense.'') to 5 (``Really
good''). They were also asked for qualitative information, such as how
the jokes might be improved, and if they had heard any of the jokes
before.

This testing was {\em not} meant to be statistically
rigorous. However, when it comes to analyzing the data, this lack of
rigour causes some problems. Because there were so few jokes and joke
judges, the scores are not statistically significant.
Moreover, there was no control group of jokes. We suspect
that jokes of this genre are not very funny even when they are produced
by humans; however, we do not know how human-produced jokes would fare
if judged in the same way \jape's jokes were, so it is difficult to make
the comparison. Ideally, with hindsight, \jape's  jokes would then have
been mixed with similar jokes (from \cite{CAJB}, for example), and then
all the jokes would have been judged by a group of schoolchildren, who
would be less likely to have heard the jokes before and more likely to
appreciate them.

\input{epsf}
\begin{figure}[htb]
\mbox{\epsfxsize=3in
\epsffile{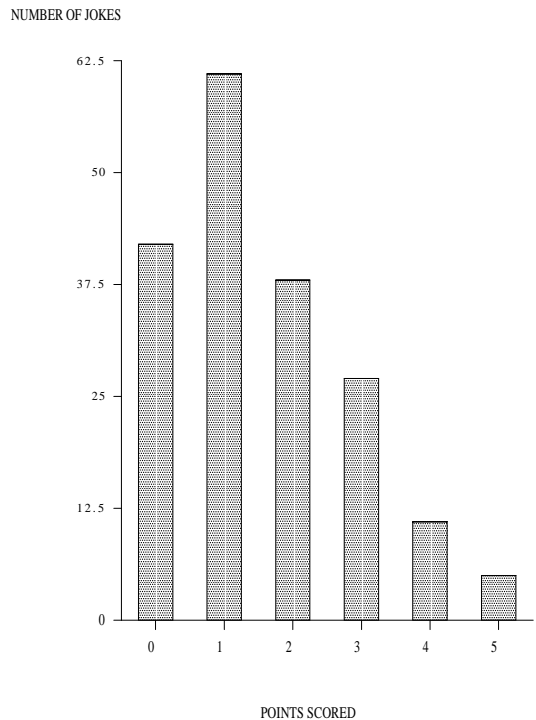}}
\caption{The point distribution over all the output}
\label{pointdis}
\end{figure}

The results of the testing are summarised in  Figure~\ref{pointdis}. The
average point score for all the jokes \jape\ produced from the lexical
data provided by volunteers is 1.5 points, over a total of 188 jokes.
Most of the jokes were given a score of 1. Interestingly, all of the
nine jokes that were given the maximum score of five by one judge, were
given low scores by the other judge --- three got zeroes, three got
ones, and three got twos. Overall, the current version of \jape\
produced, according to the scores the judges gave, ``jokes, but pathetic
ones''.  The top end of the output are definitely of Crack-a-Joke book
quality, and some (according to the judges) existed already as jokes,
including:
\begin{quote}
What do you call a murderer that has fibre? {\em A cereal killer.} \\
What kind of tree can you wear? {\em A fir coat.}\\
What kind of rain brings presents? {\em A bridal shower.}\\
What do you call a good-looking taxi? {\em A handsome cab.}\\
What do you call a perforated relic? {\em A holey grail.}\\
What kind of pig can you ignore at a party? {\em A wild bore.}\\
What kind of emotion has bits? {\em A love byte.}
\end{quote}

It was clear from the evaluation that some schemata and templates tended
to produce better jokes than others. For example, the {\em use\_syn}
template produced several texts that were judged to be non-jokes, such
as:

\begin{quote}
What do you use to hit a waiting line? {\em A pool queue.}
\end{quote}

The problem with this template is probably that it uses the definition
constructed by the schema inappropriately. The schema-generated
definition is `nonsense', in that it describes something that doesn't
exist; nonetheless, the word order of the punchline does contain some
semantic information (i.e. which of its words is the object and which
word describes that object), and it is important for the question to
reflect that information. A more appropriate template, {\em
class\_has\_rev}, produced this joke:

\begin{quote}
What kind of line has sixteen balls? {\em A pool queue.}
\end{quote}

which the judges gave an average of two points.

Another problem was that the definitions provided by the volunteers were
often too general for our purposes. For example, the entry for the word
``hanger'' gave its class as {\bf device}, producing jokes like:

\begin{quote}
What kind of device has wings? {\em An aeroplane hanger.}
\end{quote}

which scored half a point.

\section{Conclusions}
\label{improvements}

This evaluation has accomplished two things. It has shown that \jape\
can
produce pieces of text that are recognizably jokes (if not very good
ones) from a relatively unbiased lexicon. More importantly, it has
suggested some ways that \jape\ could be improved:
\begin{quote}
$\bullet$ The description of the lexicon could be made more precise, so
that
it is easier for people unfamiliar with \jape\ to make appropriate
entries.
Moreover, multiple versions of an entry could be compared for `common
knowledge', and that common knowledge entered in the lexicon.\\
$\bullet$ More slots could be added to the lexicon, allowing the person
entering words to specify what a thing is made of, what it uses, and/or
what it is part of.\\
$\bullet$ New, more detailed templates could be added, such as ones
which
would allow more complex punchlines.\\
$\bullet$ Templates and schemata that give consistently poor results
could be removed.\\
$\bullet$ The remaining templates could be adjusted so that they use the
lexical data more gracefully, by providing the right amount of
information in the question part of the riddle.\\
$\bullet$ Schema-template links that give consistently poor results
could be removed.\\
$\bullet$ \jape\ could be extended to handle other joke types, such as
simple spoonerisms and sub-word puns.
\end{quote}
If even the simplest of the trimming and ordering heuristics described
above were implemented, \jape's output would be restricted to
good--quality punning riddles. Although there is certainly room for
improvement in \jape's performance, it does produce recognizable jokes
in accordance with a model of punning riddles, which has not been done
successfully by any other program we know of. In that, it is a success.

\section{Acknowledgments}

We would like to thank Salvatore Attardo for letting us have access to
his unpublished work, and for his comments on the research reported
here.

\nocite{*}
\bibliographystyle{aaai}
\bibliography{shortbib}
\end{document}